\documentclass[10pt,twocolumn,letterpaper]{article}
\PassOptionsToPackage{numbers,sort&compress}{natbib}
\usepackage{wrapfig}
\usepackage[]{cvpr}
\usepackage{float}
\usepackage{dblfloatfix}

\usepackage{graphicx}
\usepackage{amsmath}
\usepackage{amssymb}
\usepackage{booktabs}
\usepackage{microtype}
\usepackage{placeins}
\usepackage{tabularx,array,makecell}

\definecolor{cvprblue}{rgb}{0.21,0.49,0.74}
\usepackage[pagebackref,breaklinks,colorlinks,allcolors=cvprblue]{hyperref}

\def\paperID{****}
\def\confName{CVPR}
\def\confYear{2026}

\title{TraversalBench: Challenging Paths to Follow for Vision Language Models}

\author{
Clara Petrova\\
Massachusetts Institute of Technology, Department of Physics\\
NSF AI Institute for Artificial Intelligence and Fundamental Interactions\\
{\tt\small clara\_m@mit.edu}
\and
Zhuo Chen\\
Massachusetts Institute of Technology, Department of Physics\\
Massachusetts Institute of Technology, Institute for Data, Systems, and Society\\
NSF AI Institute for Artificial Intelligence and Fundamental Interactions
\and
Marin Solja\v{c}i\'c\\
Massachusetts Institute of Technology, Department of Physics
}

\begin{document}
\maketitle

\vskip 0.3in

\begin{abstract}
Vision-language models (VLMs) perform strongly on multimodal benchmarks, but their ability to follow complex visual paths remains under-tested. We introduce \textsc{TraversalBench}, a controlled benchmark for exact visual path traversal. Each instance contains a continuous polyline with a unique start marker and labeled vertices; models must recover the ordered sequence encountered from start to finish. The benchmark balances self-intersection count, tortuosity, vertex count, and nearby confounding lines while limiting reliance on OCR, world knowledge, or open-ended planning.

We find that self-intersections are the dominant source of difficulty. A first-crossing analysis localizes failures to crossing points: performance is stable before the first crossing, then drops sharply when the model must resolve the correct continuation. Nearby confounders have weaker but compounding effects, and an auxiliary reading-order benchmark reveals a consistent left-to-right bias. Together, these results characterize how VLMs perceive and fail on visual paths. Finally, we position \textsc{TraversalBench} as a new contribution to the growing line of sustained visual grounding benchmarks for VLMs.
Code, benchmark data, and rendered examples are available at
\url{https://github.com/clarapetrova/traversalbench}.
\begin{figure*}[!h]
    \centering
    \includegraphics[width=\textwidth]{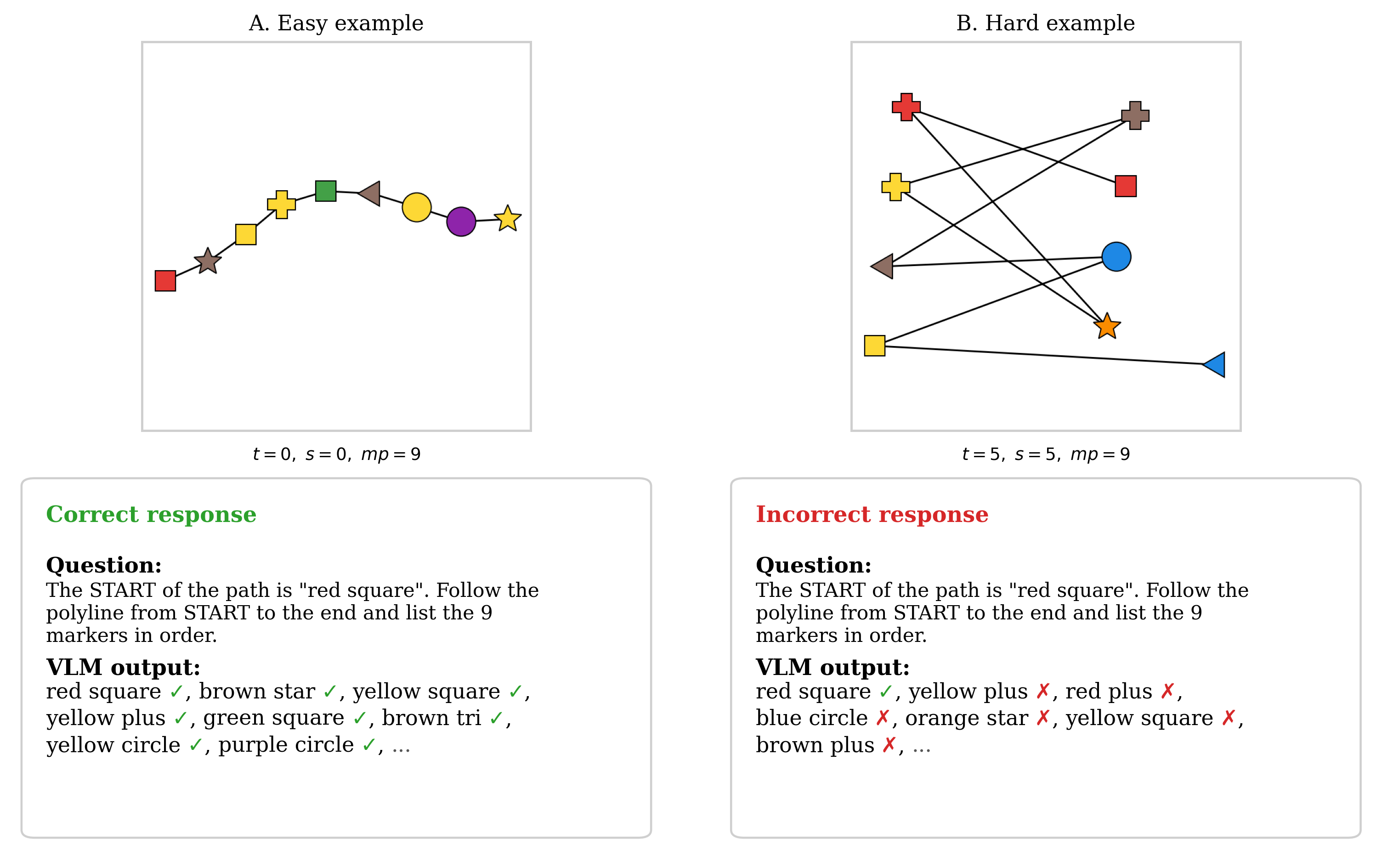}
\caption{Examples from \textsc{TraversalBench}. Left: A non-tortuous, non-self-intersecting path that the model traces correctly. Right: A highly tortuous, heavily self-intersecting path for which the model produces an incorrect sequence. Glyphs are rendered slightly larger than in examples in the dataset.}
    \label{fig:teaser}
\end{figure*}
  
\end{abstract}

\section{Introduction}

Vision-language models (VLMs) have made rapid progress on multimodal benchmarks, yet their performance remains uneven on tasks requiring sustained visual grounding. In many downstream settings, a system must trace visual structure faithfully rather than recognize it only in aggregate. For human observers, following a single continuous visual path is often straightforward\footnote{Interested readers can try a small set of representative examples in Appendix \ref{app:selftest}.}; for current models, it remains surprisingly brittle.

Recent work shows that skills needed to interpret maps, transit diagrams, and other graph-like visuals remain challenging for current models, while even simple geometric judgments such as overlaps and intersections can be surprisingly unreliable \cite{xing2025mapbench,feng2025reasonmap,bauer2025visualgraphqa,rahmanzadehgervi2024blind,berman2025vlmstunnelvisionevaluating}. More broadly, failures on tasks such as counting, localization, and shape-versus-texture perception suggest continuing limitations in binding, serial attention, and the faithful use of visual structure \cite{gavrikov2025talkmodelsseeingworld,campbell2025understandinglimitsvisionlanguage,kajic2022numbers,wu2025symmetricalvisualcontrastiveoptimization,acharya2019tallyqa,chen2025knotsimpleminimalisticenvironment,song2025visualpuzzles}. Together, these results suggest that strong aggregate benchmark performance may overstate models' ability to reason faithfully over structured visual inputs.

Our goal is not total realism per se, but the isolation of a core visual primitive that is heavily entangled in open-world tasks: maintaining the correct image-grounded continuation of a target structure over time. Existing  evaluations often entangle several sources of challenge at once: OCR, semantic interpretation of labels, domain knowledge, route planning, and free-form answer generation \cite{xing2025mapbench,feng2025reasonmap,bauer2025visualgraphqa}. Conversely, controlled reasoning benchmarks have shown the value of minimizing domain knowledge in order to isolate the capability of interest \cite{song2025visualpuzzles,acharya2019tallyqa,paiss2023teachingclip,bauer2025visualgraphqa,arnold2026mapqa}. See Appendix \ref{sec:background} for additional related work.

However, prior work does not simultaneously  test whether a model can faithfully traverse a visually presented path in exact order \emph{and} systematically balance the path-structural factors that are likely to govern traversal difficulty.
\begin{figure*}[t]
    \centering
    \includegraphics[width=\textwidth]{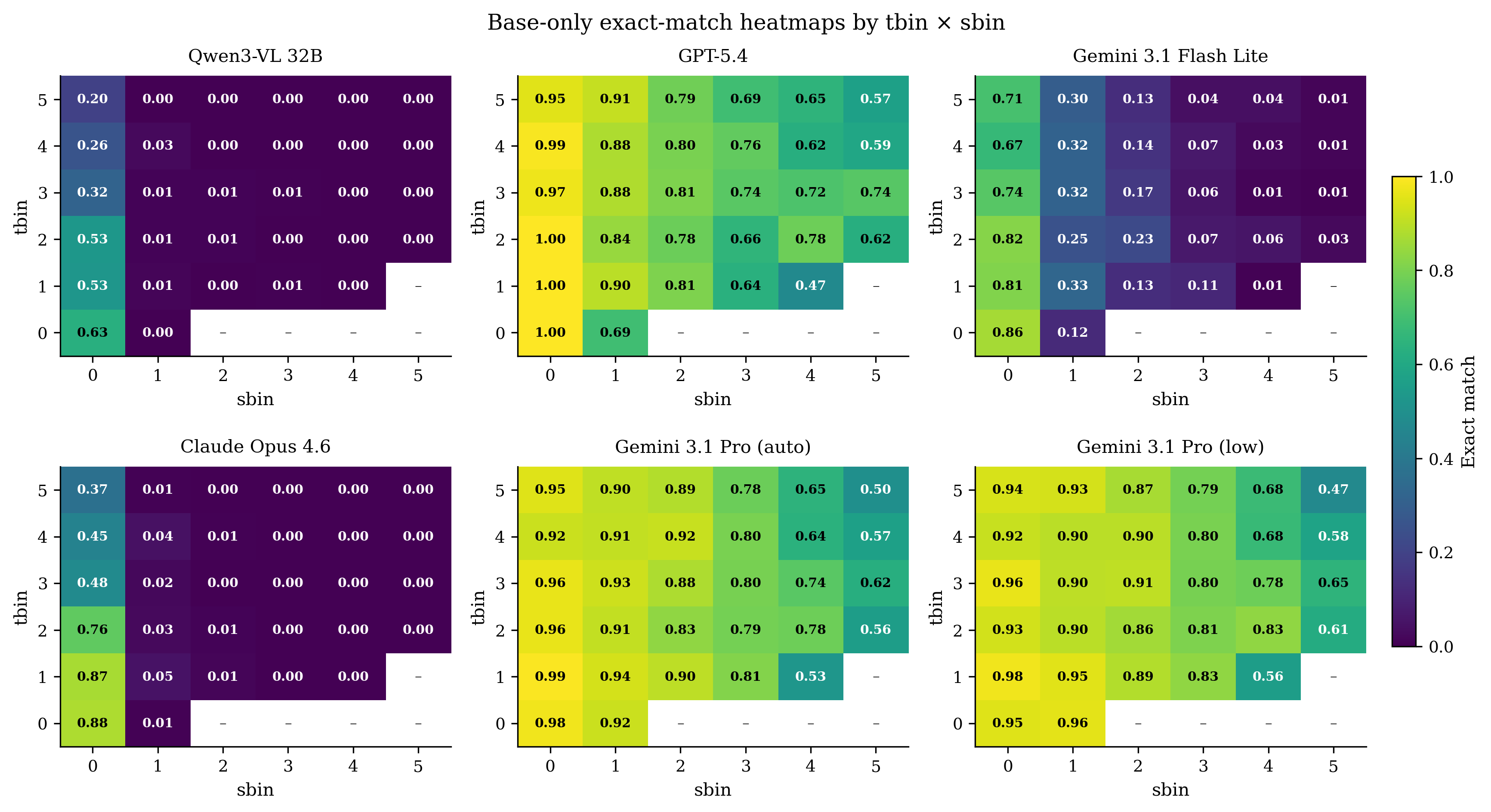}
    \caption{Exact-match accuracy across the joint grid of tortuosity and self-intersection bins for all evaluated models (non-confound). Performance is highest in the low-tortuosity, low-crossing regime and declines sharply as self-intersections increase, with tortuosity adding a further gradient of difficulty.}
    \label{fig:complexity_heatmap_em}
\end{figure*}
\begin{figure*}[b]
    \centering
    \includegraphics[width=\textwidth]{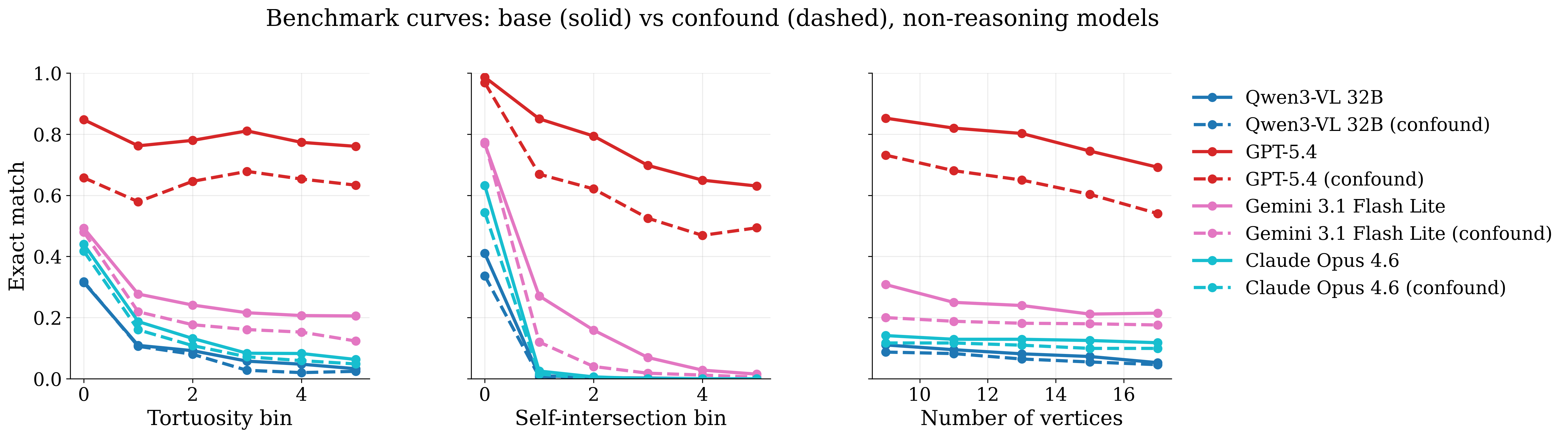}
    \caption{Benchmark curves across the main controlled factors for the non-reasoning model set. Exact-match accuracy declines as paths become more complex, with self-intersections producing the strongest degradation. Dashed curves indicate matched confound conditions, showing an additional penalty from nearby non-target structure.}
    \label{fig:benchmark_curves}
\end{figure*}
Our central question is not only whether VLMs can trace paths, but \emph{what specifically breaks them}. Using \textsc{TraversalBench}, we find that self-intersections are the strongest driver of failure, that crossings induce a sharp local continuation bottleneck, and that nearby confounding lines impose a broader and more persistent distractor burden. We also find that stronger frontier models perform well in aggregate but still remain far from saturated on exact traversal.

To summarize our contributions: 

\begin{itemize}
    \item We introduce \textsc{TraversalBench}, a controlled benchmark for exact path traversal which requires sustained visual attention.
    \item We balance four interpretable complexity factors: self-intersections, tortuosity, vertex count, and confounding lines.
    \item We show that self-intersections are the dominant source of error and that first-crossing failures are sharply localized.
    \item We show that confounding lines induce a penalty that is slightly weaker and is primarily cumulative.
    \item We provide an auxiliary analysis of reasoning budget, finding that additional reasoning can help on difficult examples but does not consistently improve end-task success, and may come at the cost of reliability and latency.
\end{itemize}

\section{Benchmark Design and Dataset Construction}

\subsection{Task and Controlled Factors}

Our benchmark isolates path-faithful visual reasoning in a knowledge-light setting. Each instance contains a single continuous path, a unique start token, and a target sequence defined by the ordered markers encountered along the path. This minimizes reliance on OCR, world knowledge, and open-ended planning, allowing us to focus on visual traversal itself.

To support fine-grained analysis, we explicitly vary four interpretable axes of difficulty: tortuosity, self-intersection count, number of vertices, and confounding lines. 

Here, tortuosity denotes how winding a path is, quantified as the ratio of total path length to end-to-end displacement; higher values indicate more circuitous paths.
Tortuosity is discretized using thresholds $\{1.0, 1.3, 2.0, 3.0, 4.5, 6.5, 9.0\}$, yielding six bins: $[1.0,1.3)$, $[1.3,2.0)$, $[2.0,3.0)$, $[3.0,4.5)$, $[4.5,6.5)$, and $[6.5,9.0)$. The lower bins are loosely motivated by values discussed in prior work on natural meandering systems \cite{bledsoe2001logistic,leopold1957river}, while the higher bins deliberately extend as we progress into more extreme regimes in order to stress model performance beyond naturally common path shapes.

Self-intersection count is discretized using thresholds $\{0,1,2,4,6,9,13\}$, corresponding to the regimes $0$, $1$, $2$--$3$, $4$--$5$, $6$--$8$, and $9$--$12$ crossings. Here the motivation is less about matching a canonical natural taxonomy and more about visual traceability: prior work on graph readability and transit-map design consistently identifies edge crossings and crossing geometry as important determinants of readability and path-following difficulty \cite{purchase1997aesthetics,huang2010crossingangle,wu2020transitmap, ware2002cognitive}. We therefore use especially fine granularity near zero to separate uncrossed, singly crossed, and lightly woven paths, while grouping the dense high-crossing regime more coarsely.

Motivated by human work on attentional contour tracing and target–distractor segregation \cite{houtkamp2003gradual}, we also add confounding lines onto existing backbones. Confounding lines are nearby non-target segments added to increase local distractor structure without changing the target path itself.

Representative examples illustrating these controlled factors are provided in Appendix Fig~\ref{fig:complexity_examples}.

\subsection{Generation Overview}

We generate polyline backbones to target discrete cells in a joint tortuosity/self-intersection grid, then render each accepted backbone with glyphs placed at vertices. Acceptance is render-aware: candidates must satisfy geometric constraints on segment length, nonlocal clearance, and vertex-to-segment separation, ensuring that benchmark difficulty is driven primarily by controlled path structure rather than accidental rendering artifacts. See Appendix \ref{sec:si_constraints} for further details.

\subsection{Dataset Summary}

\begin{figure}[t]
    \centering
    \includegraphics[width=\linewidth]{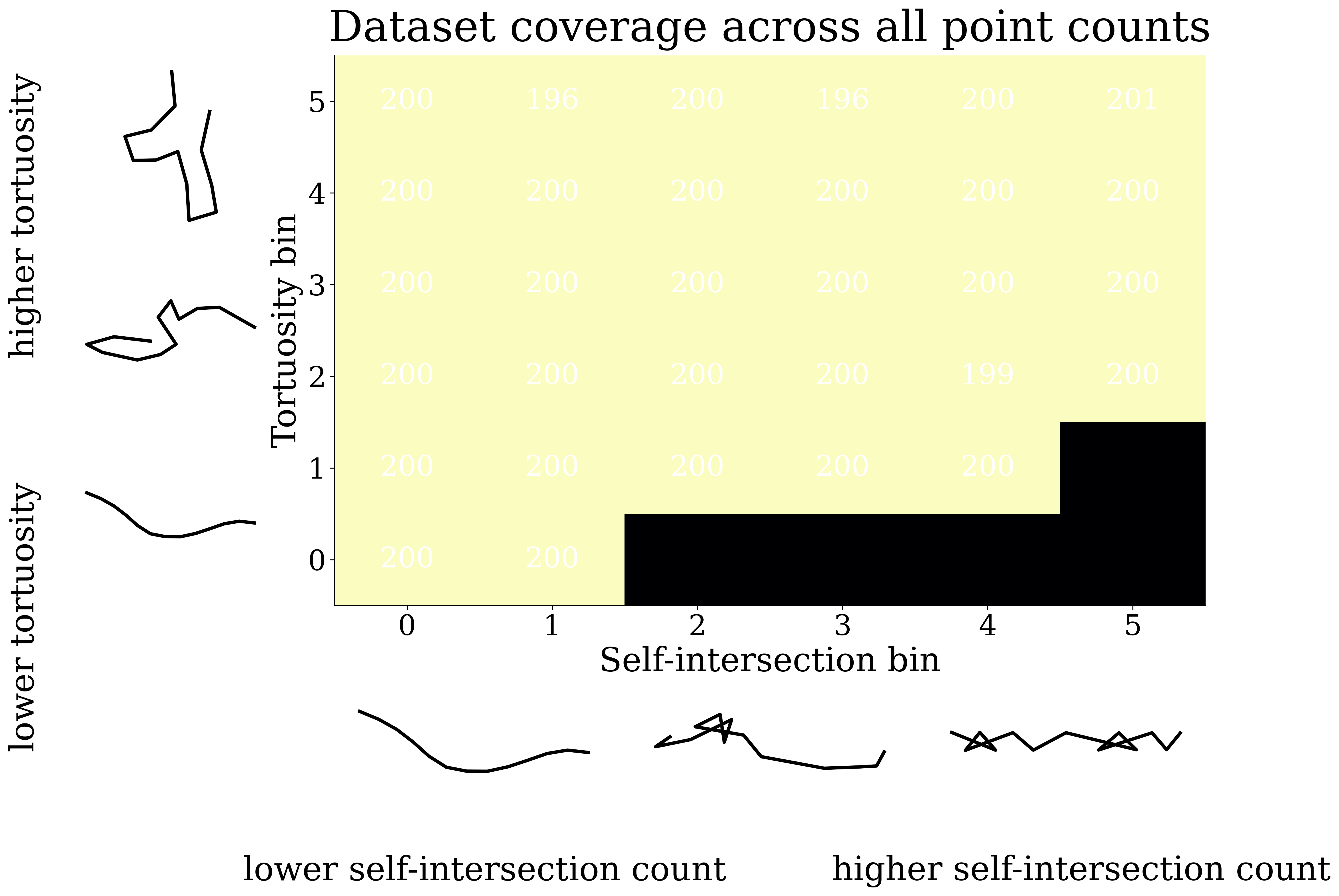}
    \caption{Coverage of the new-model  benchmark across tortuosity and self-intersection bins, pooled over point counts. Example backbone curves placed to the left and below the heatmap illustrate the progression from lower to higher tortuosity and self-intersection without the visual clutter of glyph rendering. Covered cells are in yellow, uncovered cells are in black.}
    \label{fig:new_model_coverage_heatmap}
\end{figure}

The resulting benchmark is balanced across populated complexity cells and multiple point-count settings. This design allows us to separate the effects of crossings, windingness, path length, and confounding structure rather than collapsing them into a single undifferentiated notion of difficulty. In the base (non-confound) dataset, there are 6192 unique examples. In the confound dataset, there are 6200 unique examples. Please see Figure~\ref{fig:new_model_coverage_heatmap} for details; not all cells can be covered due to constraints described in 2.2 and in Appendix \ref{sec:si_constraints}. 
\begin{figure*}[b]
    \centering
    \includegraphics[width=\textwidth]{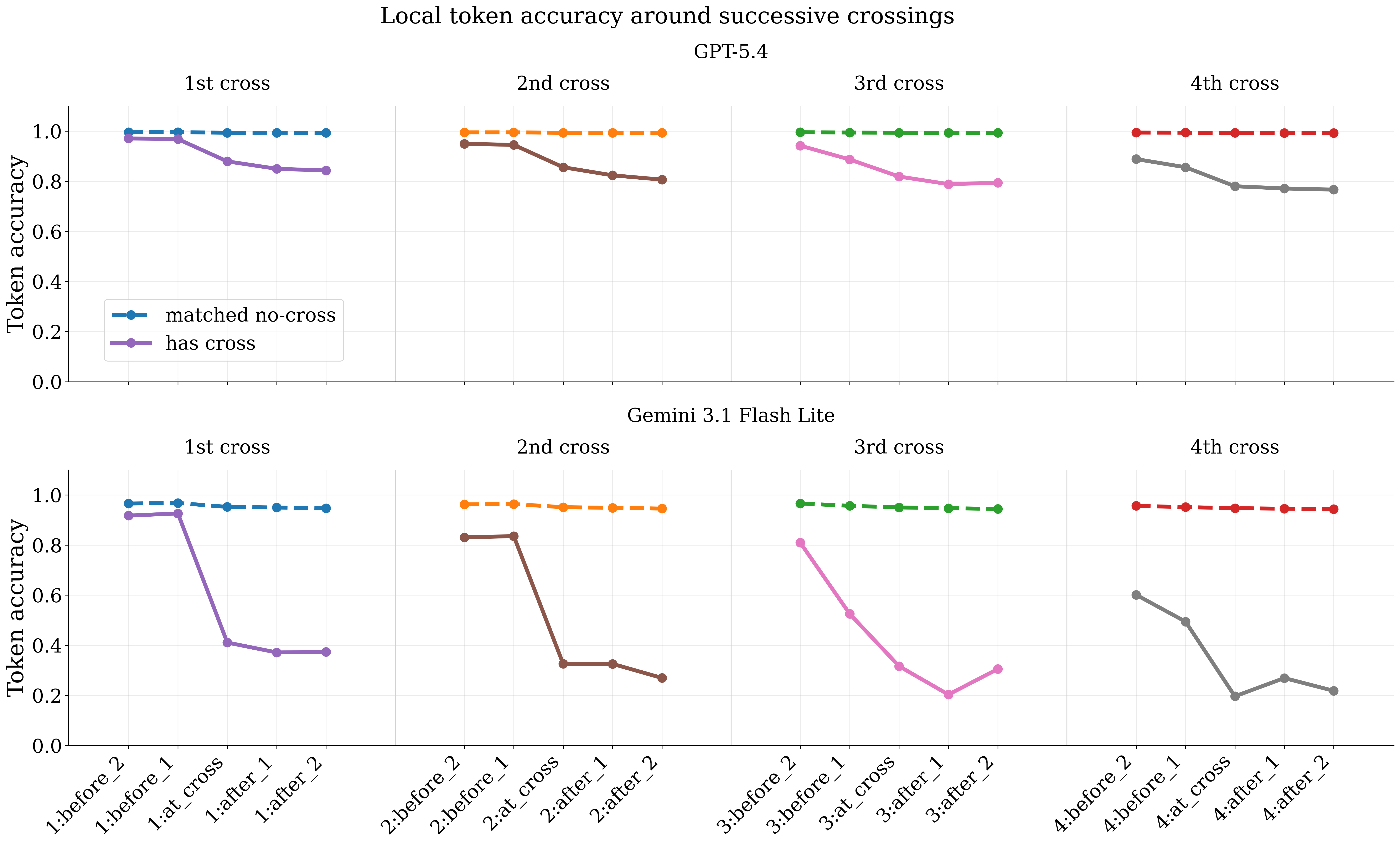}
    \caption{Local token accuracy around the first through fourth crossings. Accuracy is relatively stable immediately before the crossing, then drops sharply at the crossing itself, consistent with a localized continuation-selection bottleneck. Dashed curves show position-matched no-cross controls.}
    \label{fig:first_crossing_window}
\end{figure*}

\section{Experiments and Results}

\subsection{Experimental Setup}

We evaluate six vision-language models spanning stronger and weaker proprietary and open-weight systems: Gemini 3.1 Pro (low reasoning), Gemini 3.1 Pro (auto reasoning), GPT-5.4, Gemini 3.1 Flash Lite, Claude Opus 4.6, and Qwen3-VL 32B. We use a shared prompt template and a shared output-format constraint across models (see Appendix \ref{sec:si_prompting}). We report exact-match accuracy (EM), which requires the full predicted sequence to match the ground truth exactly, and token-level accuracy (TokAcc), computed positionwise after standardized parsing.

\subsection{Main Results}

Exact visual path traversal remains challenging across all evaluated models. On the base benchmark, Gemini 3.1 Pro (low) performs best at 0.823 EM / 0.949 TokAcc, followed by Gemini 3.1 Pro (auto) at 0.814 / 0.922 and GPT-5.4 at 0.782 / 0.891. Performance then drops sharply to Gemini 3.1 Flash Lite at 0.245 / 0.626, Claude Opus 4.6 at 0.128 / 0.465, and Qwen3-VL 32B at 0.082 / 0.419.

A large gap remains between exact match and token-level accuracy across models. This indicates that many failures are not wholly random outputs, but structured traversal errors: models often recover part of the correct sequence before making a localized mistake.

We also find that Gemini 3.1 Pro (low) slightly outperforms Gemini 3.1 Pro (auto) while using substantially fewer reasoning tokens, suggesting that more internal reasoning does not simply translate into better exact traversal.

\subsection{Performance by Path Complexity}

Performance declines systematically as paths become more complex. Figure~\ref{fig:benchmark_curves} summarizes the marginal performance trends across tortuosity, self-intersection count, and vertex count, with dashed curves showing matched confound conditions.

 As the strongest and clearest degradation occurs along the self-intersection axis, this indicates that crossings are the dominant controlled source of difficulty. Tortuosity and vertex count contribute additional difficulty, but more gradually.

The same broad pattern appears in the joint tortuosity/self-intersection grid. Figure~\ref{fig:complexity_heatmap_em} visualizes exact-match accuracy across this grid for all evaluated models, showing that the hardest regime is the jointly high-tortuosity, high-self-intersection region.

Confounding lines impose an additional penalty beyond the base geometric trend. This suggests that traversal difficulty is not determined only by the target path's own topology: nearby non-target structure can also interfere with faithful tracing.
\begin{figure*}[t]
    \centering
    \includegraphics[width=\textwidth]{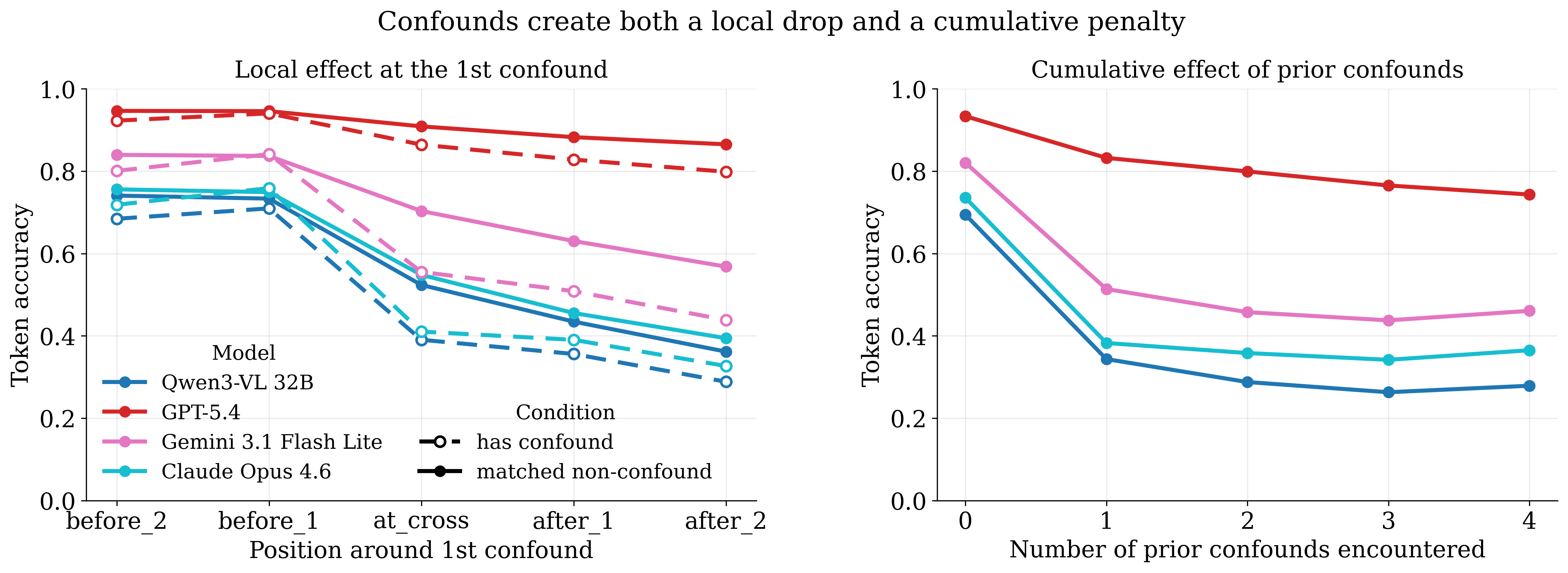}
    \caption{Effects of nearby confounding lines on traversal performance. Left: local token accuracy around the first confound, with solid curves showing matched non-confound controls. Right: token-level accuracy as a function of the number of previously encountered confounds. Confounds produce a primarily cumulative degradation across generated tokens.}
    \label{fig:confound_effects}
\end{figure*}
\section{Difficulty Analysis}

\subsection{Self-Intersections Create a Sharp Local Bottleneck}

Self-intersections are the most damaging controlled factor in the benchmark. Relative to non-intersecting paths, all models show large declines in both exact-match and token-level accuracy. A first-crossing analysis shows that this effect is not merely global: performance is often relatively strong immediately before the first crossing, then drops sharply at the crossing itself when the model must choose the correct continuation. For stronger models, this appears as a localized cliff superimposed on otherwise good prefix performance; for weaker models, the drop is much larger and often followed by sustained drift. This pattern suggests that many failures arise not from a complete inability to follow a path sequentially, but from instability at local ambiguity points where multiple continuations are visually plausible. Rather than behaving as though they cannot trace at all, models often appear able to maintain the target path over substantial prefixes and then fail at the moment when renewed visual disambiguation is required. This is broadly consistent with prior graph-readability and transit-map work identifying crossings as a natural source of path-following difficulty for human readers as well \cite{purchase1997aesthetics,huang2010crossingangle,wu2020transitmap}.

Thus, self-intersections impose two separable burdens: paths that eventually self-intersect are already somewhat harder even before the first crossing, but the crossing itself also creates a distinct local continuation bottleneck.
\begin{figure*}[b]
    \centering
    \includegraphics[width=\linewidth,height=0.34\textheight,keepaspectratio]{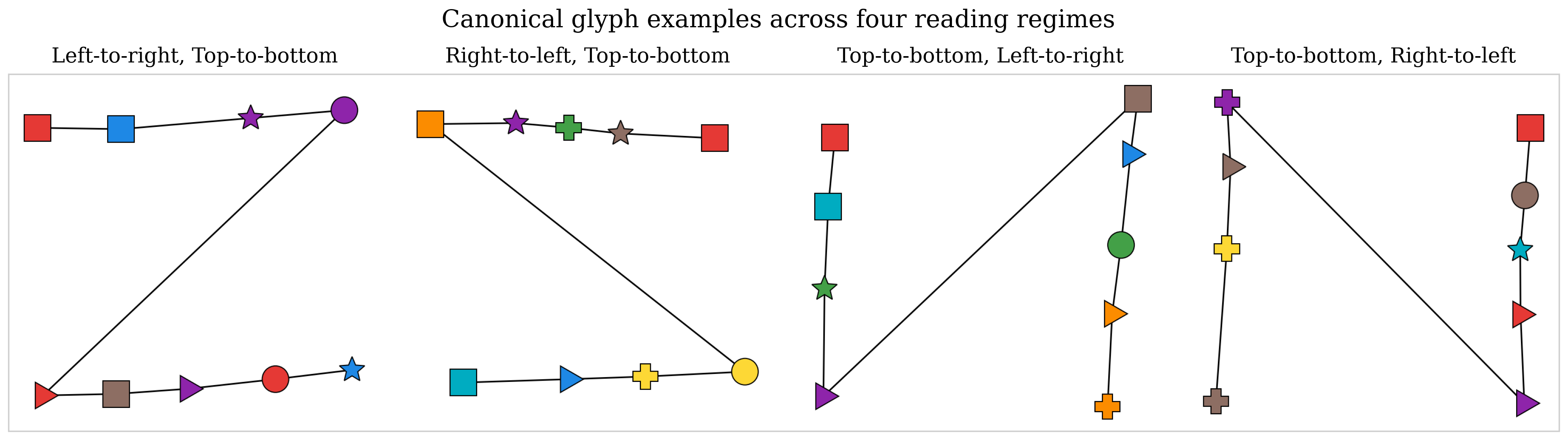}
    \caption{Representative examples from the four reading-order regimes used in the auxiliary analysis. The panels illustrate the different regimes for 9 vertices and 2 lanes; glyphs are rendered slightly larger than actual dataset entries.}
    \label{fig:reading_regimes}
\end{figure*}

\subsection{Confounding Lines Induce Persistent Interference}

Nearby confounding lines also impair traversal, but differently from self-intersections. Whereas crossings produce a sharp local drop at a continuation decision, confounds induce a broader and more persistent degradation. Performance on confounded examples remains below matched non-confound controls, indicating that the added lines actively interfere with faithful traversal rather than merely correlating with harder examples, as shown in Figure~\ref{fig:confound_effects}. Although weaker in magnitude than crossings, this effect compounds with repeated exposure: as the number of previously encountered confounds increases, token-level accuracy continues to decline. These results suggest that confounds impose a cumulative distractor burden over the sequence, rather than a purely local ambiguity event.

\begin{figure*}[t]
    \centering
    \includegraphics[width=\textwidth]{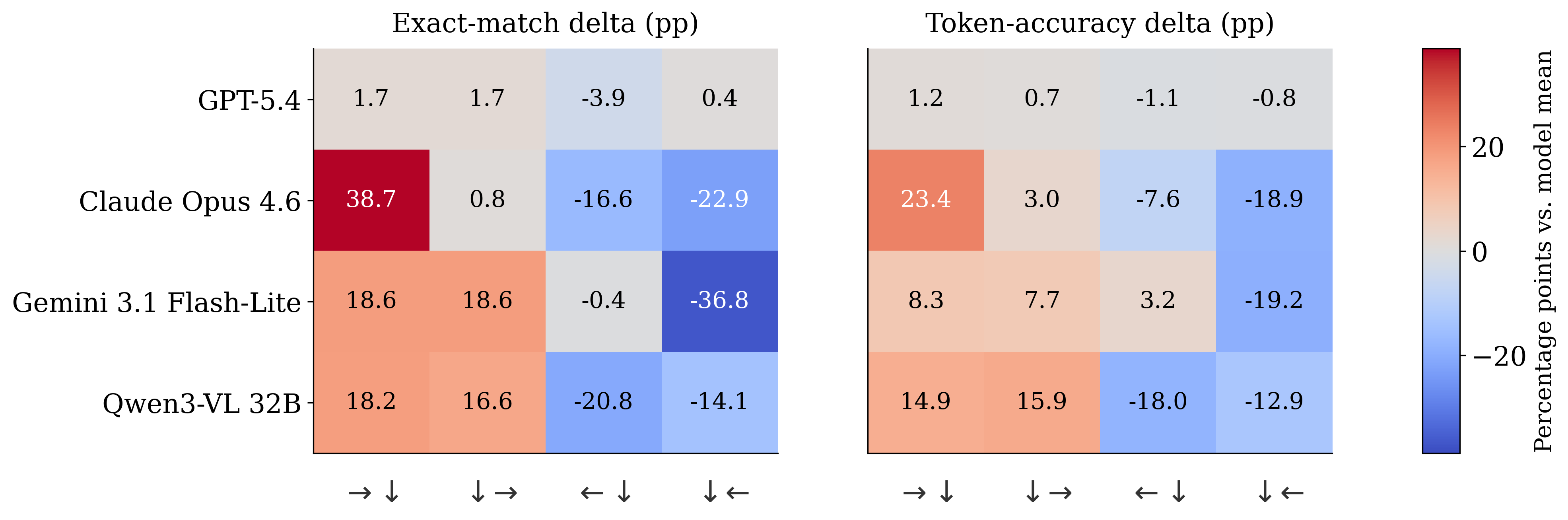}
    \caption{Regime effects on the auxiliary reading-order benchmark by model. Each cell shows the change in performance, in percentage points, relative to that model's own average across regimes. Positive values indicate easier-than-average regimes for that model, and negative values indicate harder-than-average regimes. Across models, \texttt{ltr\_tb} $\rightarrow \downarrow$ and \texttt{ttb\_lr} $\downarrow \rightarrow$ are generally easier, while \texttt{rtl\_tb} $\leftarrow \downarrow$ and especially \texttt{ttb\_rl} $\downarrow \leftarrow$ are harder.}
    \label{fig:language_regimes}
\end{figure*}
\begin{table}
\centering
\caption{Auxiliary reading-order benchmark results for the representative models. TokAcc = token-level accuracy. The final two columns report the estimated change in token accuracy, in percentage points, from the minimum to maximum observed lane count and vertex count, respectively, based on model-specific regressions controlling for regime.}
\label{tab:language_results_aux}
\begin{tabular}{lrrrr}
\toprule
Model & TokAcc & \shortstack{$\Delta$TokAcc\\(lanes)} & \shortstack{$\Delta$TokAcc\\(vertices)} \\
\midrule
              GPT-5.4 & 98.7\% &        +0.3 pts &           -0.9 pts \\
      Claude Opus 4.6 & 63.9\% &       -21.0 pts &           -7.5 pts \\
Gemini 3.1 Flash-Lite & 90.1\% &        -7.8 pts &           -4.9 pts \\
         Qwen3-VL 32B & 67.8\% &       -21.3 pts &           -6.3 pts \\
\bottomrule
\end{tabular}
\end{table}
\subsection{Auxiliary Analysis: Reading-Order Bias}

We include this auxiliary benchmark to test whether learned layout priors could partially confound main-benchmark performance, not to replace the main geometric analysis. This dataset contains four explicitly structured path regimes: left-to-right then top-to-bottom (\texttt{ltr\_tb}), right-to-left then top-to-bottom (\texttt{rtl\_tb}), top-to-bottom then right-to-left (\texttt{ttb\_rl}), and top-to-bottom then left-to-right (\texttt{ttb\_lr}). Representative examples from these four regimes are shown in Figure~\ref{fig:reading_regimes}. Each regime is crossed with lane count and point count under the same view-space and glyph-aware rendering constraints as the main generator, yielding a controlled test of whether exact traversal is easier when path layout resembles familiar scan-order structure.

We evaluate the same model set used in the main benchmark. To isolate directional and layout preferences from simple difficulty effects, we macro-average performance across all $(\texttt{lane\_count}, \texttt{n\_points})$ combinations within each regime. Overall results on this auxiliary benchmark are reported in Table~\ref{tab:language_results_aux}. To highlight regime-specific preferences independent of overall model strength, Figure~\ref{fig:language_regimes} shows the change in performance, in percentage points, relative to each model's own mean across regimes.

\subsection{Reasoning Budget Does Not Reliably Rescue Traversal}
\begin{table}[!h]
\centering
\small
\setlength{\tabcolsep}{4pt}

\begin{minipage}{0.98\linewidth}
\centering
\textbf{A. Gemini 3.1 Pro reasoning settings}\\[2pt]
\begin{tabular}{lccc}
\toprule
Reasoning setting & TokAcc &     EM & Mean reasoning tokens \\
\midrule
              low & 94.9\% & 82.3\% &                   663 \\
             auto & 92.2\% & 81.4\% &                  1527 \\
\bottomrule
\end{tabular}

\end{minipage}

\vspace{0.6em}

\begin{minipage}{0.92\linewidth}
\centering
\textbf{B. GPT-5.4 vs GPT-5.4 Pro overlap subset}\\[2pt]
\begin{tabular}{lcc}
\toprule
               Metric & GPT-5.4 & GPT-5.4 Pro \\
\midrule
                   Exact match &  47.3\% &      57.5\% \\
          Answer rate & 100.0\% &      60.4\% \\
   EM $\mid$ answered &  47.3\% &      95.2\% \\
Mean reasoning tokens &       0 &        2334 \\
\bottomrule
\end{tabular}

\end{minipage}

\caption{Reasoning can improve traversal, but not reliably. \textbf{Top:} Gemini 3.1 Pro low slightly outperforms auto on the base benchmark while using far fewer reasoning tokens. \textbf{Bottom:} On the overlap subset (hard examples with $tbin, sbin \geq 4$ and no confounders) covered by GPT-5.4 Pro, the reasoning-enabled variant improves exact match over GPT-5.4 overall, but its answer rate is substantially lower; when it does return an answer, it is usually correct.}
\label{tab:reasoning_summary}
\end{table}

Additional reasoning can help on exact visual traversal, but not in a clean or uniformly beneficial way. On a hard matched overlap subset, a reasoning-enabled GPT-5.4 Pro variant improves over GPT-5.4 overall, indicating that extra inference-time reasoning can improve performance on visually demanding examples. At the same time, its answer rate is substantially lower: when GPT-5.4 Pro does return a final sequence, it is usually correct, but it often fails to produce one at all. This suggests that the remaining gap is driven partly by budget-related failure rather than incorrect answered outputs, while also leaving open the possibility that performance would eventually saturate even with more room to reason. Reasoning therefore appears to improve capability in one sense while making final-task success less reliable. See Appendix \ref{app:gpt54pro_harder_answered}.

The Gemini 3.1 Pro comparison points in a related but distinct direction. Here, the lower-reasoning setting slightly outperforms the automatic setting while using far fewer reasoning tokens overall. This suggests that additional internal reasoning is not a universal remedy for exact traversal, and strengthens the possibility that simply increasing reasoning budget will not eliminate the remaining errors. Instead, reasoning appears to help only when it remains effectively coupled to visual grounding; otherwise, longer internal deliberation may become inefficient or even counterproductive. (Table~\ref{tab:reasoning_summary}).

\section{Discussion}

Exact visual path traversal remains a substantial blind spot for current VLMs. Even the strongest evaluated models remain far from saturated on exact sequence recovery. The main failure mode is not uniform collapse, but localized breakdown at key structural bottlenecks.

These findings also sharpen the relationship between our results and recent tunnel-vision accounts of VLM failure \cite{berman2025vlmstunnelvisionevaluating}. Prior work argues that current VLMs struggle with nonlocal visual reasoning, including smooth visual search along continuous contours. Our results are consistent with that broad concern, but they suggest a more specific failure mode than a blanket inability to follow paths. Many models appear able to trace substantial segments of the paths correctly and then fail at crossings or with repeated confound exposure. We therefore interpret the bottleneck less as pure tunnel vision than as unstable visual grounding during sequential tracing.

 Recent work suggests that visual information in multimodal language models is mediated by a relatively limited subset of attention heads \cite{Bi_2025_CVPR}, that models can devote substantial attention mass to irrelevant visual sink tokens \cite{kang2025see}, that visual-token attention in deeper layers can become highly inefficient \cite{chen2024image}, and that visual attention can weaken and become noisier as responses lengthen \cite{jung2025visual}. Taken together, these results are consistent with a failure mode in which models can often trace substantial prefixes correctly, but do not sufficiently re-anchor to the image when local ambiguity arises. On this view, additional reasoning helps only insofar as it is paired with renewed visual grounding. In real-world settings, this also matters operationally: increasing reasoning time may increase latency without reliably delivering the answer quality one actually needs \emph{on the timescale }one needs it.

A central takeaway is that not all sources of difficulty operate in the same way. Self-intersections create sharp local continuation failures, while confounding lines induce a broader distractor burden. Controlled benchmarks are useful precisely because they allow these failure modes to be separated rather than collapsed into a single aggregate score. We anticipate that in the future we will add further controls for curvature (a local rather than global parameter) and widen the dataset into harder regimes.

It remains unclear what would most effectively remediate these failures. Because benchmarks for sustained visual attention in multimodal models are still relatively sparse, it is difficult to tell whether the deficits we identify here can be substantially reduced by straightforward finetuning alone. Our results suggest that simply increasing inference-time reasoning is unlikely to be a complete solution. A more plausible path may involve substantially more training data of the relevant kind, combined with training procedures that more strongly encourage sustained image-grounded computation over the course of generation. At the same time, our results do not by themselves imply that the primary bottleneck is architectural. Further research is likely necessary to disambiguate the ultimate cause of these failures.

In summary, we present \textsc{TraversalBench} as a useful diagnostic for path-faithful visual reasoning, a tool for tracking progress on fine-grained spatial grounding amidst ambiguity and clutter, and a contribution to the growing field of sustained visual attention tasks.
\paragraph{Social impact.}
The primary contribution of this work is methodological: a benchmark for measuring path-following and spatial reasoning performance in vision-language models. We do not anticipate substantial direct societal harms from the benchmark itself, though stronger evaluation may support safer and more transparent use of these systems.

\FloatBarrier
\clearpage

\bibliographystyle{ieeenat_fullname}

\bibliography{cites}
\newpage
\appendix
\onecolumn
\section{Supplementary Methods}
\label{sec:si_methods}

\subsection{Backbone generation overview}
\label{sec:si_backbone}

\begin{figure}[t]
    \centering
    \includegraphics[width=\linewidth]{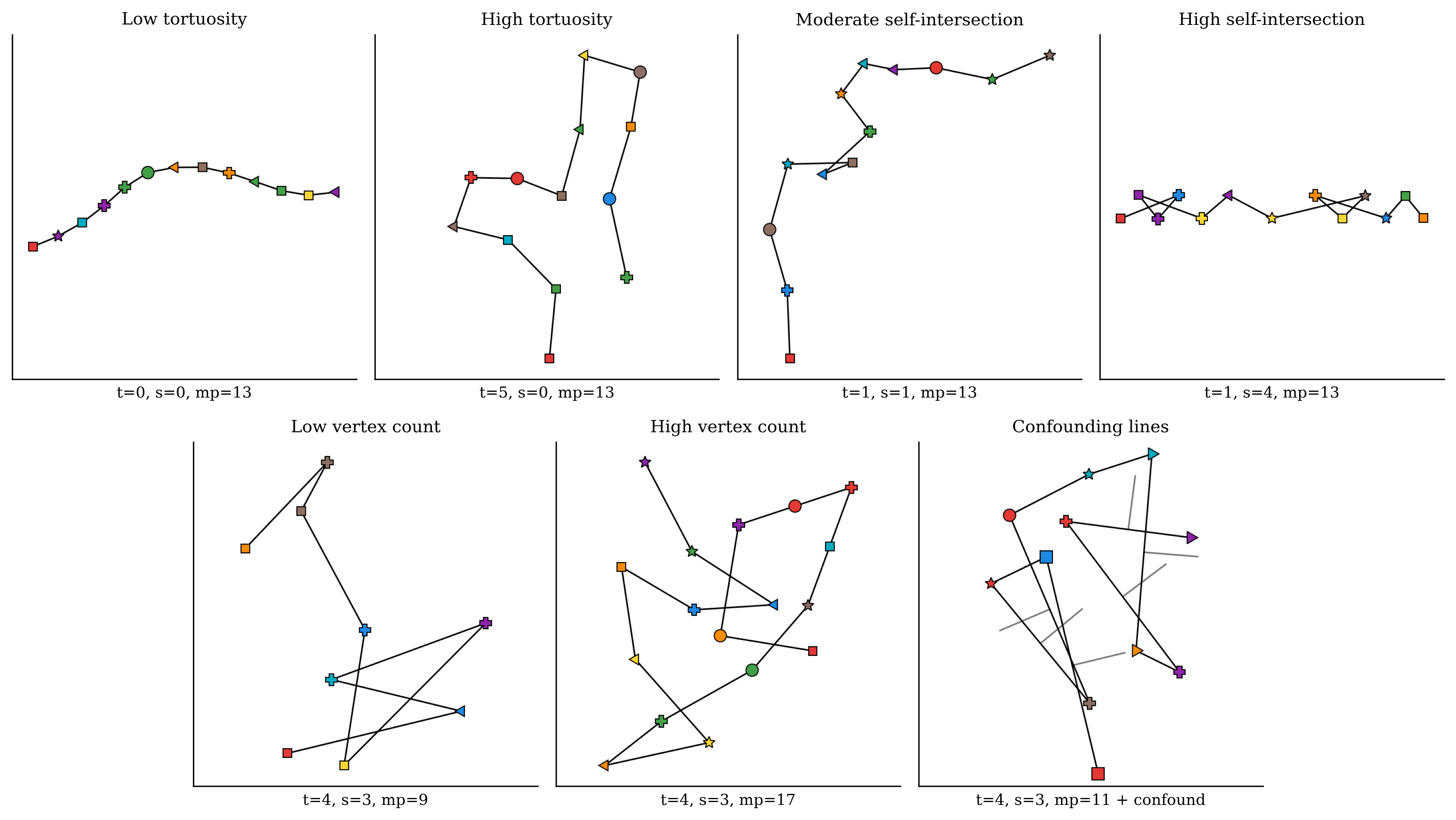}
    \caption{Representative benchmark instances illustrating the main controlled axes of path complexity. The panels show variation in tortuosity, self-intersection structure, vertex count, and the presence of confounding lines. Tortuosity increases as the path becomes more winding, self-intersection increases as the path crosses itself more often, lower versus higher vertex counts corresponds to fewer versus more vertices, and confounding lines add nearby distractor structure without changing the target path.}
    \label{fig:complexity_examples}
\end{figure}
We generate unlabeled polyline ``backbones'' to target discrete cells in a joint grid of tortuosity and self-intersection count. Tortuosity is binned using thresholds $\{1.0, 1.3, 2.0, 3.0, 4.5, 6.5, 9.0\}$, yielding six bins, and self-intersection count is binned using thresholds $\{0,1,2,4,6,9,13\}$, yielding the regimes $0$, $1$, $2$--$3$, $4$--$5$, $6$--$8$, and $9$--$12$. A small set of cells is explicitly skipped because they were treated as unattainable under the generation constraints. Each accepted record is written as JSONL with an integer id, a generator-family label, the vertex list, measured metrics, and target metadata. The script is configured around a default square view of 672 px, chosen to work well with AnyRes-style 336-px tiling, and includes optional tile-seam avoidance to reduce crop-boundary failures in downstream VLM evaluation; the latter option was defaulted to off as it limited the number of acceptable backbones to an extreme degree. 
\subsection{Render-aware geometric constraints}
\label{sec:si_constraints}

Acceptance is render-aware rather than purely topological. In addition to matching the requested tortuosity and self-intersection bins, candidates must pass a suite of view-space constraints after fitting into the final render window. These include a minimum segment length, a minimum overall extent, nonlocal segment--segment separation, vertex-to-nonincident-segment separation, and a hard cap on near-reversal angles. The generator is also glyph-aware: when vertex glyphs are enabled, the script derives stricter effective spacing thresholds from the glyph radius and padding so that rendered markers do not overlap. Concretely, if $r$ is the glyph radius and $p$ is the glyph padding, the script defines a minimum center-to-center spacing $S = 2r + p + 10$ px, raises both the minimum segment length and the minimum nonlocal vertex spacing to at least $S$, requires non-intersecting segments to remain separated by at least $2r$, and enlarges the intersection-masking radius so that true crossings are not spuriously rejected by dense-gap checks. Optional seam padding further keeps vertices away from AnyRes crop seams.

\subsection{Family-specific generators and sparse-cell handling}

The generator combines a broad set of family-specific constructions with more generic fallback families. Low- and mid-complexity cells are often populated using structured constructions such as bowties, rail-weaves, split stars, braided or knot-like templates, and proposal-gated path builders. Harder cells may be reached by template deformation, controlled permutation of crossing structure, anisotropic scaling, local warping, endpoint extension, or repeated safe splitting under the same render-aware acceptance gate. The code also explicitly distinguishes families whose point sets should be preserved from families that can be resampled. For a few sparse high-difficulty cells, the script supports a bootstrap mode: it loads previously accepted backbones from a source run at lower point count, converts them to normalized world-like coordinates, refits them into the current view, and then grows them to the target vertex count by safe view-space insertions. In the current configuration, this mechanism is enabled especially for $(t,s)=(3,3)$ and $(3,4)$ when generating 17-point examples.

\subsection{Near-duplicate filtering}
Near-duplicate filtering is done within each $(t,s)$ cell using a canonical polyline signature and a signature-distance threshold, with a bounded number of representative shapes retained per cell. This keeps the final dataset both valid and morphologically diverse, while still allowing deterministic multi-pass generation and recovery from interrupted runs. 

\subsection{Evaluation details}
\label{sec:si_eval}
Unless otherwise noted, all models were evaluated using default inference settings. For proprietary models accessed via OpenRouter, we used the default decoding configuration, with low verbosity and automatic image detail enabled where available. We used medium reasoning for GPT-5.4 Pro, while we used low for Gemini 3.1 Pro  (low), and auto for Gemini 3.1 Pro (auto). For Qwen3-VL 32B, we likewise used default inference settings. We did not use chain-of-thought-style prompting or model-specific prompt engineering beyond the common prompt template and output-format constraints described below.

\subsection{Prompting and output format}
\label{sec:si_prompting}

\begin{table*}[t]
\centering
\small
\setlength{\tabcolsep}{4pt}
\caption{Matched-prefix control analysis at the first crossing. For each intersecting example, we compare exact correctness on the prefix up to the first crossing against a no-cross baseline matched by prefix length. Negative deltas indicate that prefixes on paths that eventually self-intersect are already harder than equally long prefixes on non-intersecting paths. N = 4992.}
\label{tab:first_crossing}
\begin{tabular}{lllll}
\toprule
             Model & Intersecting prefix exact & Matched no cross prefix exact & Delta vs matched no cross & Mean pre len \\
\midrule

Qwen3-VL 32B & 72.9\% & 82.7\% & -9.8 pts & 3.14 \\
GPT-5.4 & 96.6\% & 99.5\% & -2.9 pts & 3.14 \\
Gemini 3.1 Flash Lite & 90.4\% & 94.0\% & -3.6 pts & 3.14 \\
Claude Opus 4.6 & 74.9\% & 87.1\% & -12.1 pts & 3.14 \\
Gemini 3.1 Pro (auto) & 96.1\% & 98.8\% & -2.8 pts & 3.14 \\
Gemini 3.1 Pro (low) & 98.5\% & 98.8\% & -0.3 pts & 3.14 \\
\bottomrule
\end{tabular}
\end{table*}
We used a single task framing across models and varied only the output format during pilot experiments. The shared instruction stated that the image contained a single continuous polyline with colored shape markers at each vertex, that there were no branches, and that exactly one next marker followed from each step along the path. Evaluation under the confounder version of the dataset used the same prompt, but specified that gray spurs should be ignored and that the target line should be followed. The prompt identified a unique start token by its color--shape combination and instructed the model to traverse the path continuously from that start to the endpoint, listing all encountered markers in order.

In early pilots, we compared three output formats built on this same instruction: (i) a comma-separated list, (ii) a JSON array of strings, and (iii) a numbered line-by-line list. Although all three formats were parseable in principle, the comma-separated format was the most robust across models and also yielded the best overall accuracy. We therefore used it for all main experiments.

The final user prompt further tightened the format in several ways. First, it explicitly required the model to output exactly \(N\) markers, where \(N\) was instantiated from the ground-truth path length for that example. This reduced premature stopping and under-generation although it lends help to the model (worse performance, including time-outs, was observed when removing the "exactly  \(N\)..." stipulation). Second, it specified the exact lexical form of each item as \texttt{<color> <shape>} in lowercase. Third, it enumerated the allowed color and shape vocabularies, which reduced spelling and ontology drift. Finally, it instructed the model to reply only as a comma-separated list of exactly \(N\) items, with no extra words or markdown.

We paired this with a short system instruction requesting strict format compliance:
\begin{quote}
\small
\texttt{You are a strict evaluation function. Output ONLY the final answer in the exact format requested. No preamble, no explanation, no coordinates, no bullets, no extra lines, no markdown/code fences.}
\end{quote}

The final user prompt template was:
\begin{quote}
\small
\texttt{You are shown a single polyline path with colored shape markers at each vertex. The path is a single continuous polyline (no branches). There is exactly one next marker at each step.}

\texttt{The START of the path is the marker "\{START\_TOKEN\}". This exact combo appears only once. Starting from START, follow the polyline continuously and list every marker you encounter in order until the path ends.}

\texttt{You must output EXACTLY \{N\} markers. If you are uncertain, still give your best guess---do not stop early.}

\texttt{Each marker must be written exactly as: <color> <shape> (lowercase). Allowed colors: red, blue, green, orange, yellow, cyan, purple, brown. Allowed shapes: circle, square, tri, star, plus.}

\texttt{Reply ONLY as a comma-separated list of exactly \{N\} items. No extra words. No markdown/code fences.}
\end{quote}

\section{Additional Results for the Main Benchmark}

\subsection{Joint-grid token-accuracy heatmap}
\begin{figure}[t]
    \centering
    \includegraphics[width=\linewidth]{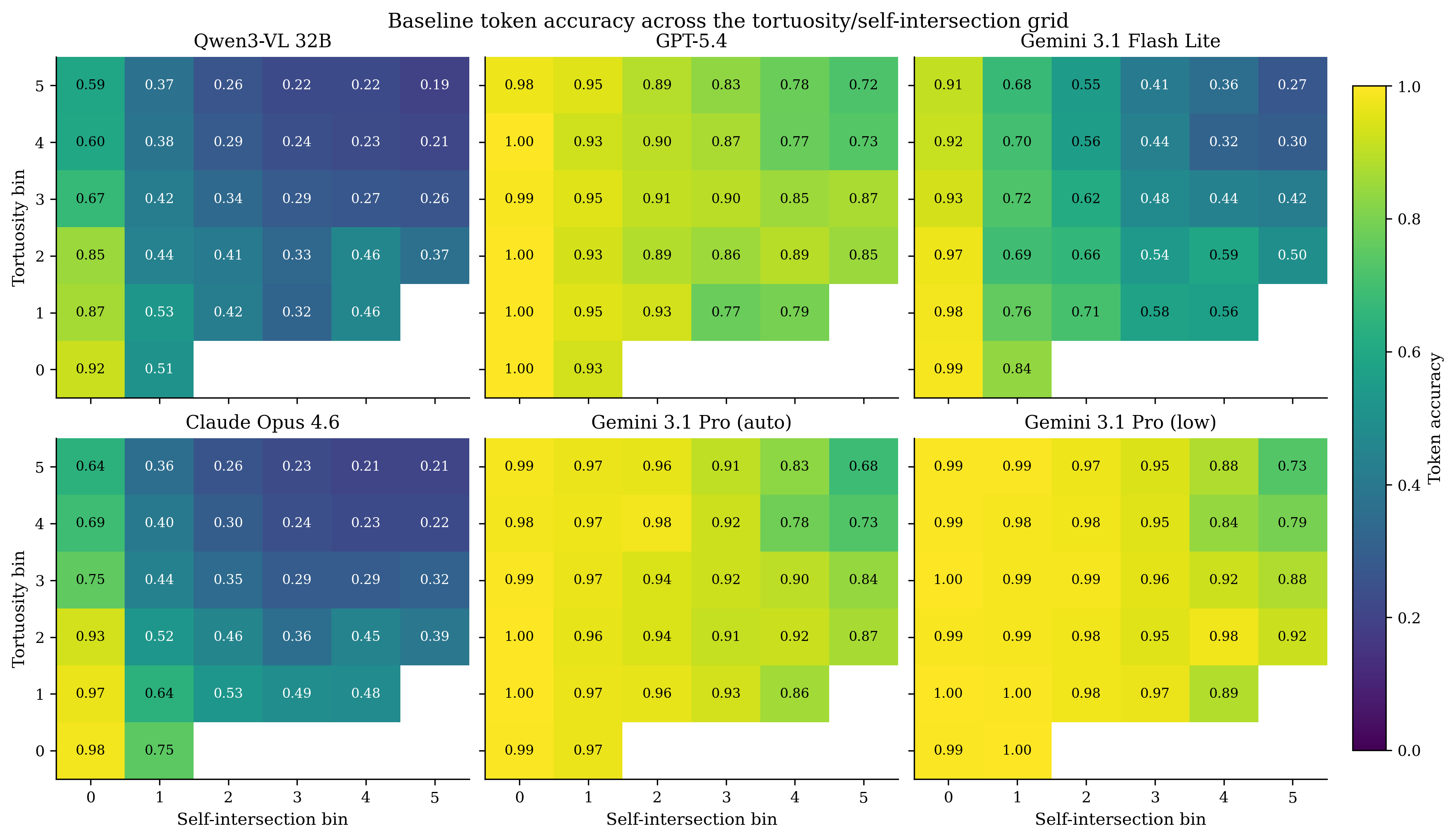}
    \caption{Token accuracy across the joint grid of tortuosity and self-intersection bins for each model on the base benchmark. Performance is highest in the low-tortuosity, low-crossing regime and declines sharply as self-intersections increase, with tortuosity adding a further gradient of difficulty.}
    \label{fig:complexity_heatmap}
\end{figure}

Here we include other results from the main benchmark, including token accuracies for each model. Since exact matches are not very sparse, we relegate this result to the appendix. Please see Figure~\ref{fig:complexity_heatmap}.

\subsection{Self-crossings globally increase difficulty}
We include a table describing how token accuracy is affected before a crossing; paths that self-intersect are less likely to be correct even before the crossing occurs. Our results suggest self-crossings introduce a global difficulty increase apart from the local effects described in the main text.

\subsection{GPT-5.4 Pro reasoning heatmaps on harder examples}
\label{app:gpt54pro_harder_answered}

To better interpret GPT-5.4 Pro on the harder examples, we separate overall exact match from answer rate and exact match conditioned on producing an answer. This distinction matters because, on the hardest cells, GPT-5.4 Pro often fails by not returning a final answer at all, so raw exact match conflates abstention with conditional answer quality.

On the exact overlap subset covered by the GPT-5.4 Pro run (480 examples spanning 12 $(mp,s,t)$ cells, with $mp \in \{13,15,17\}$), GPT-5.4 achieves 47.3\% exact match, while GPT-5.4 Pro achieves 57.5\% exact match overall. However, GPT-5.4 Pro produces a final answer on only 60.4\% of examples. Conditional on answering, its exact-match rate rises to 95.2\%. Thus, GPT-5.4 Pro's failures on this subset are driven much more by answer suppression than by low conditional accuracy once an answer is produced.

This pattern becomes even clearer as difficulty increases. At $mp=13$, GPT-5.4 Pro answers 84.4\% of examples and reaches 97.0\% exact match conditional on answering. At $mp=15$, the answer rate drops to 51.9\%, while exact match conditional on answering remains 92.8\%. At $mp=17$, the answer rate falls further to 45.0\%, but exact match conditional on answering is still 94.4\%. The same qualitative pattern appears at the cell level: in the hardest cells, GPT-5.4 Pro often answers only a small fraction of examples, yet when it does answer, it is usually correct. For example, at $(mp,s,t)=(17,5,5)$, the model answers only 12.5\% of examples, but all interpretation of the raw heatmaps should therefore keep answer rate in view. The hardest regions remain challenging, but much of the raw degradation reflects the model's tendency not to emit a final answer, rather than uniformly poor performance on answered cases. It is unclear how many tokens would be necessary for the model to complete reasoning on these problems, and whether the answer would be correct in these cases. 

\begin{figure*}[t]
    \centering
    \includegraphics[width=\textwidth]{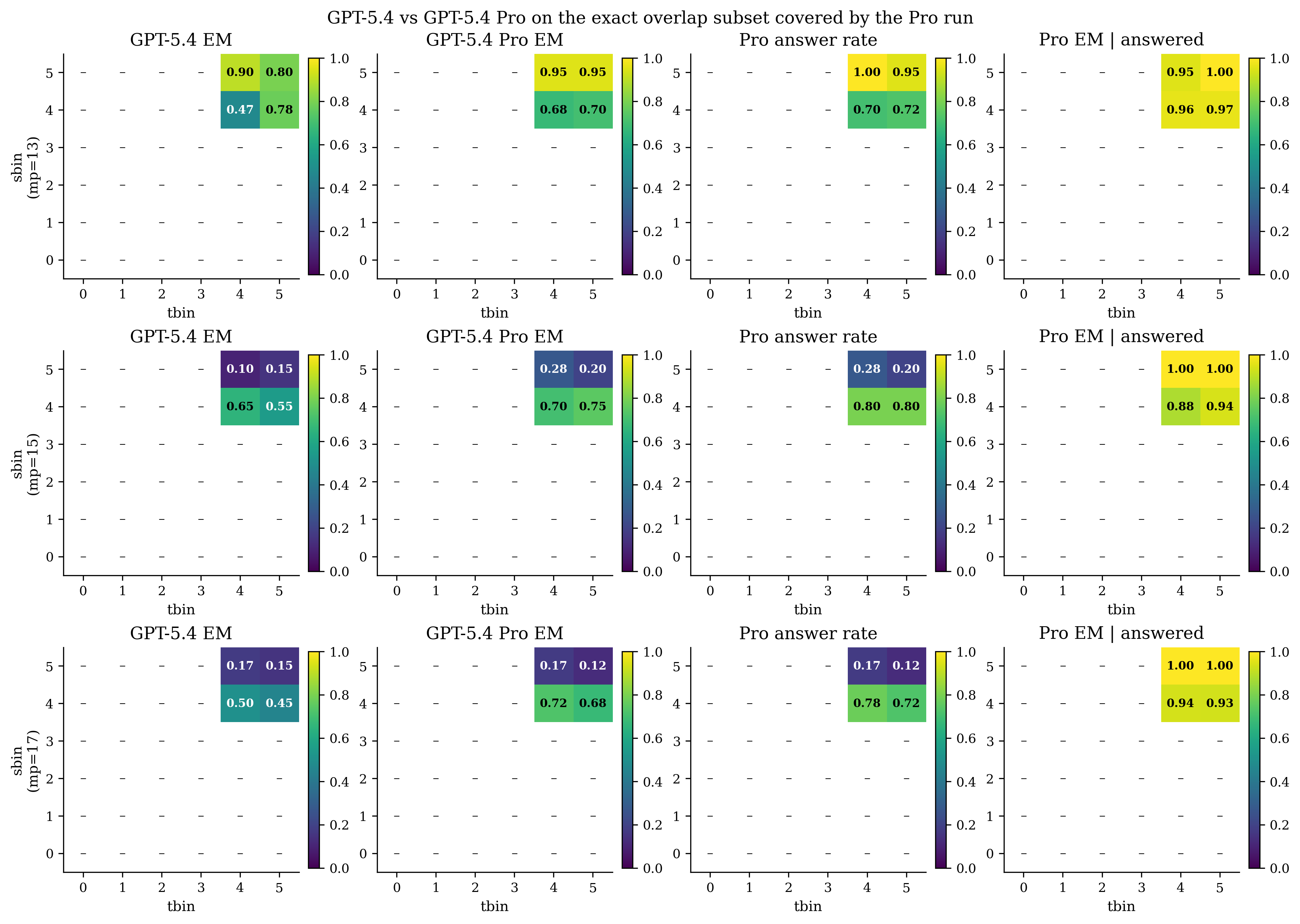}
    \caption{GPT-5.4 versus GPT-5.4 Pro on the exact overlap subset covered by the GPT-5.4 Pro run. The panels show GPT-5.4 exact match, GPT-5.4 Pro exact match, GPT-5.4 Pro answer rate, and GPT-5.4 Pro exact match conditioned on answering, across the hard-example cells. GPT-5.4 Pro's degradation on the hardest cells is driven in large part by failures to emit a final answer; conditional on answering, accuracy remains high.}
    \label{fig:gpt54pro_overlap_heatmaps}
\end{figure*}

\subsection{Control for coarse left-to-right orientation}
\label{sec:si_lr_control}
As a control, we tested whether the main benchmark's difficulty trends could be explained by a coarse left-to-right (LR) versus right-to-left (RL) directional bias, defined by the net horizontal displacement from the designated start to the endpoint. In pooled regressions over the new-model base benchmark, adding an LR/RL indicator leaves the main difficulty effects largely intact. The negative coefficients associated with increasing self-intersection remain large after LR/RL control: for TokAcc, the self-intersection coefficients change from -34.0 to -15.3 points before control to -33.4 to -14.4 points after control, and for EM from -49.4 to -28.9 points to -49.0 to -28.3 points. Point-count effects are also essentially unchanged (-6.0 to -1.7 points before control versus -5.7 to -1.8 after). Tortuosity effects attenuate somewhat more after LR/RL control, but remain clearly negative (-13.8 to -2.2 points before control versus -11.4 to -0.9 after). Overall, the central difficulty trends in the benchmark are not explained by coarse directional bias.

These results show that the self-intersection effect is not an artifact of averaging together easier LR examples and harder RL examples: performance declines sharply with self-intersection even after controlling for coarse directional bias.
\section{Extended Background and Related Work}
\label{sec:background}
\paragraph{Map, diagram, and graph-like reasoning in VLMs.}
Recent work shows that VLMs remain challenged by structured spatial artifacts such as maps, transit diagrams, and graph-like visuals \cite{xing2025mapbench,feng2025reasonmap,arnold2026mapqa,bauer2025visualgraphqa}. These benchmarks establish that multimodal reasoning on diagrammatic inputs is difficult, but they typically evaluate broad competence on realistic artifacts, where OCR, semantic interpretation, route planning, and free-form answer generation are entangled. Our benchmark is complementary: rather than testing general map understanding, we isolate \emph{exact path-faithful traversal} and explicitly stratify  by path-structural difficulty.

\paragraph{Knowledge-light diagnostic benchmarks.}
A parallel line of work argues that multimodal evaluation should isolate reasoning from domain knowledge and shortcut cues \cite{song2025visualpuzzles,acharya2019tallyqa,paiss2023teachingclip}; this issue frequently confounds results in prior work. Our benchmark generally fits into this philosophy by reducing the task to a unique start token, a single continuous path, and an exact ordered output sequence.

Recent controlled benchmarks have also begun to probe topological and nonlocal spatial reasoning more directly. For example, Knot So Simple studies image-based knot manipulation tasks with complexity parameterized by crossing count, highlighting the difficulty of reasoning over entangled spatial structure from visual input \cite{chen2025knotsimpleminimalisticenvironment}. Our benchmark is complementary: rather than interactive manipulation, we isolate exact sequence recovery along a single static path with controlled variation in self-intersection count, tortuosity, and distractor structure.

VisualPuzzles is especially relevant in positioning our benchmark as a knowledge-light diagnostic of visual reasoning rather than a test of familiarity with map conventions or domain semantics \cite{song2025visualpuzzles}. Controlled counting benchmarks make a similar methodological point: narrowly targeted evaluations can reveal capability gaps that broad benchmarks obscure. \cite{bauer2025visualgraphqa} is also an especially relevant work as it utilizes artificial subway-like graphs to test models. However, their work does not balance across path complexity variables, leading to a disproportionally low-tortuosity, low-self-crossing dataset. Our work explicitly balances across these measures both in the low and high regimes. 

\begin{table}[t]
\centering
\caption{Stability of the main benchmark difficulty effects after controlling for coarse LR/RL orientation. The table summarizes coefficient ranges from pooled regressions over the base benchmark. Adding an LR/RL indicator changes the main coefficients only modestly, especially for self-intersection, indicating that the central difficulty trends are not explained by directional bias.}
\label{tab:lr_control_stability_new_models}
\begin{tabular}{llll}
\toprule
                 Outcome / effect family & Before LR control & After LR control &         Change \\
\midrule
TokAcc, self-intersection ($s{=}1$--$6$) &    -34.0 to -15.3 &   -33.4 to -14.4 &   +0.6 to +1.3 \\
    EM, self-intersection ($s{=}1$--$6$) &    -49.4 to -28.9 &   -49.0 to -28.3 & +0.38 to +0.83 \\
       TokAcc, tortuosity ($t{=}1$--$5$) &     -13.8 to -2.2 &    -11.4 to -0.9 &   +1.2 to +2.7 \\
   TokAcc, point count ($mp{=}11$--$17$) &      -6.0 to -1.7 &     -5.7 to -1.8 & -0.06 to +0.39 \\
\bottomrule
\end{tabular}

\end{table}

\paragraph{Path-following difficulty: tracing, crossings, and line geometry.}
Classic work in cognitive science and graph readability suggests that path-following difficulty depends on structural properties of the path itself, including distance along the path, curvature, continuity, crossings, and local distractor structure \cite{jolicoeur1986curve,jolicoeur1991properties,houtkamp2003gradual,ullman1984visualroutines,purchase1997aesthetics,ware2002cognitive,dawson2015searchset,huang2010crossingangle,wu2020transitmap}. Curve-tracing studies motivate the view that following a connected path is a serial attentional operation rather than a trivial perceptual readout \cite{jolicoeur1986curve,jolicoeur1991properties,houtkamp2003gradual,ullman1984visualroutines}, while graph and transit-map studies identify crossings, bends, continuity, and line geometry as key determinants of readability and path-tracing performance \cite{purchase1997aesthetics,ware2002cognitive,dawson2015searchset,huang2010crossingangle,wu2020transitmap}. Our benchmark works these variables into a controlled VLM setting by balancing our artificial paths across highly interpretable axes of path complexity, most notably self-crossing count and tortuosity.

    \section{Appendix Self-Test for Interested Readers}
\label{app:selftest}

\begin{figure*}[t]
    \centering
    \includegraphics[width=\textwidth]{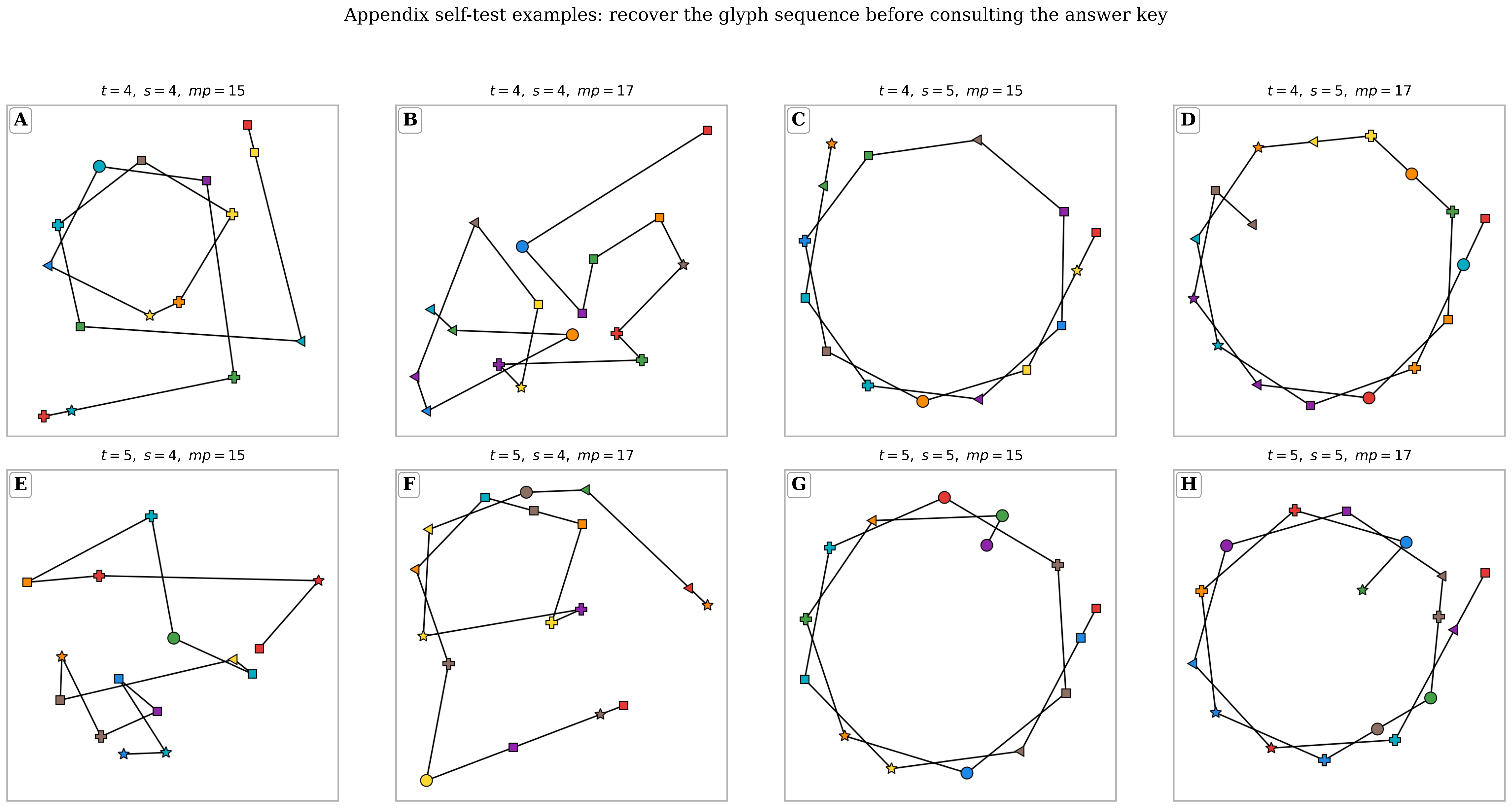}
    \caption{Appendix self-test examples for interested readers. The panels span representative high-tortuosity, high-self-intersection cases at $mp \in \{15,17\}$. Readers may try to recover the target glyph sequence before consulting Table~\ref{tab:appendix_selftest_answer_key}. These examples are included for qualitative intuition only and do not constitute a formal human study.}
    \label{fig:appendix_selftest}
\end{figure*}

To give readers an intuitive sense of the task, we include a small set of difficult representative examples (Fig~\ref{fig:appendix_selftest}) that can be attempted by eye before consulting the answer key (Table~\ref{tab:appendix_selftest_answer_key}). These examples are not intended as a formal human evaluation, and we do not report human accuracy here, although the authors themselves solved them correctly. Rather, they serve as a qualitative sanity check on the benchmark and illustrate the kinds of local ambiguity induced by tortuosity and self-intersections.

\begin{table*}[t]
\centering
\small
\setlength{\tabcolsep}{8pt}
\renewcommand{\arraystretch}{1.2}
\caption{Answer key for the appendix self-test examples.}
\label{tab:appendix_selftest_answer_key}
\begin{tabular}{c c c c p{0.58\textwidth}}
\toprule
Panel & $t$ & $s$ & $mp$ & Answer \\
\midrule
A & 4 & 4 & 15 & \texttt{red square yellow square cyan tri green square cyan plus brown square yellow plus orange plus yellow star blue tri cyan circle purple square green plus cyan star red plus} \\
B & 4 & 4 & 17 & \texttt{red square blue circle purple square green square orange square brown star red plus green plus purple plus yellow star yellow square brown tri purple tri blue tri orange circle green tri cyan tri} \\
C & 4 & 5 & 15 & \texttt{red square yellow star yellow square orange circle brown square blue plus green square brown tri purple square blue square purple tri cyan plus cyan square green tri orange star} \\
D & 4 & 5 & 17 & \texttt{red square cyan circle orange plus purple square cyan star cyan tri orange star yellow tri yellow plus orange circle green plus orange square red circle purple tri purple star brown square brown tri} \\
E & 5 & 4 & 15 & \texttt{red square red star red plus orange square cyan plus green circle cyan square yellow tri brown square orange star brown plus purple square blue square cyan star blue star} \\
F & 5 & 4 & 17 & \texttt{red square brown star purple square yellow circle brown plus orange tri cyan square brown square orange square yellow plus purple plus yellow star yellow tri brown circle green tri red tri orange star} \\
G & 5 & 5 & 15 & \texttt{red square blue square brown tri yellow star cyan square cyan plus red circle brown plus brown square blue circle orange star green plus orange tri green circle purple circle} \\
H & 5 & 5 & 17 & \texttt{red square purple tri cyan plus red star blue tri purple circle purple square brown tri brown plus green circle brown circle blue plus blue star orange plus red plus blue circle green star} \\
\bottomrule
\end{tabular}
\end{table*}

\end{document}